\documentclass[10pt,twocolumn,letterpaper]{article}

\usepackage{cvpr}
\usepackage{times}
\usepackage{epsfig}
\usepackage{graphicx}
\usepackage{amsmath}
\usepackage{amssymb}
\usepackage{bbm}
\usepackage{xcolor}
\usepackage{nth}
\usepackage{mathptmx}
\newcommand\Mark[1]{\textsuperscript{#1}}
\usepackage[toc,page]{appendix}

\usepackage[pagebackref=true,breaklinks=true,letterpaper=true,colorlinks,bookmarks=false]{hyperref}

\cvprfinalcopy % *** Uncomment this line for the final submission

\newcommand\blfootnote[1]{%
  \begingroup
  \renewcommand\thefootnote{}\footnote{#1}%
  \addtocounter{footnote}{-1}%
  \endgroup
}

\begin{document}
%%%%%%%%% TITLE
\title{Transformation-Grounded Image Generation Network\\for Novel 3D View Synthesis}

\author{Eunbyung Park\Mark{1} \quad Jimei Yang\Mark{2} \quad Ersin Yumer\Mark{2} \quad Duygu Ceylan\Mark{2} \quad Alexander C. Berg\Mark{1}\\
\Mark{1}University of North Carolina at Chapel Hill \quad \Mark{2}Adobe Research\\
{\tt\small eunbyung@cs.unc.edu \quad \{jimyang,yumer,ceylan\}@adobe.com \quad aberg@cs.unc.edu}}

\maketitle
%\thispagestyle{empty}

%%%%%%%%% ABSTRACT
\begin{abstract}
We present a transformation-grounded image generation network for novel 3D view synthesis from a single image. Instead of taking a `blank slate' approach, we first explicitly infer the parts of the geometry visible both in the input and novel views and then re-cast the remaining synthesis problem as image completion. Specifically, we both predict a flow to move the pixels from the input to the novel view along with a novel visibility map that helps deal with occulsion/disocculsion. Next, conditioned on those intermediate results, we hallucinate (infer) parts of the object invisible in the input image. In addition to the new network structure, training with a combination of adversarial and perceptual loss results in a reduction in common artifacts of novel view synthesis such as distortions and holes, while successfully generating high frequency details and preserving visual aspects of the input image. We evaluate our approach on a wide range of synthetic and real examples. Both qualitative and quantitative results show our method achieves significantly better results compared to existing methods.
\end{abstract}

%%%%%%%%% BODY TEXT
\section{Introduction}
\blfootnote{Project homepage: \url{http://www.cs.unc.edu/~eunbyung/tvsn}}
\begin{figure*}[t]
\begin{center}
\includegraphics[width=\linewidth]{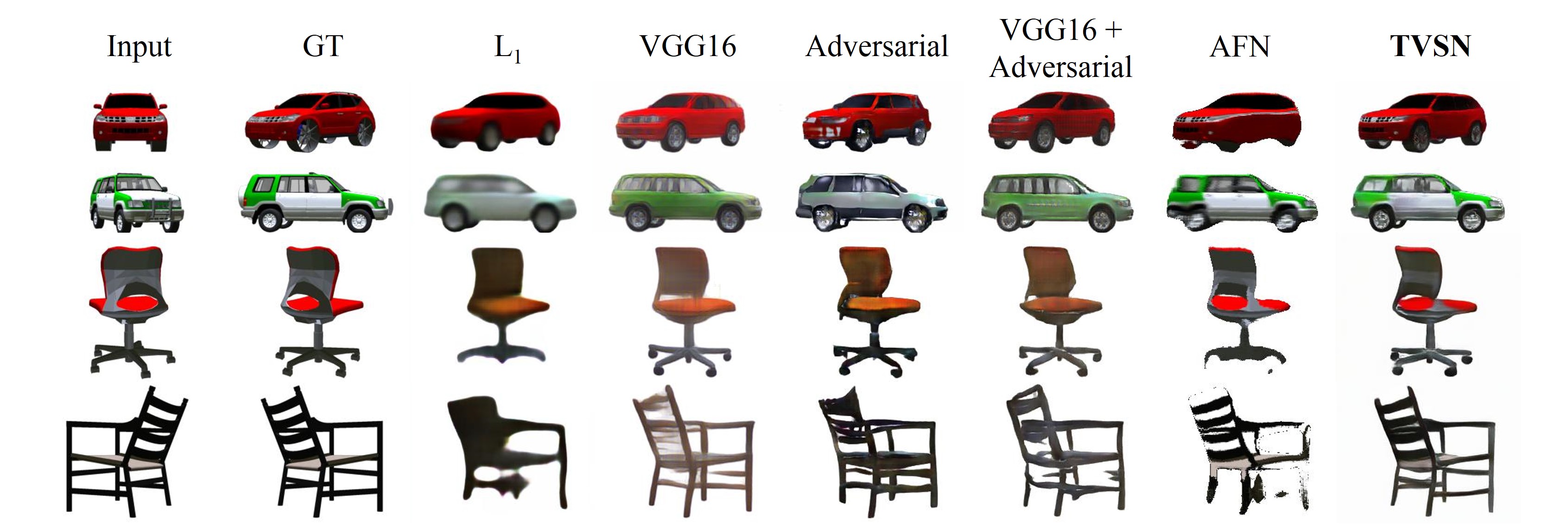}
\label{fig:teaser}
\end{center}
\caption{Results on test images from 3D ShapeNet dataset~\cite{shapenet2015}. \nth{1}-input, \nth{2}-ground truth. From \nth{3} to \nth{6} are deep encoder-decoder networks with different losses. (\nth{3}-$L_1$ norm~\cite{tatarchenko_eccv2016}, \nth{4}-feature reconstruction loss with pretrained VGG16 network~\cite{johnson_eccv2016,larsen_icml2016,ulyanov_icml2016,lamb2016discriminative}, \nth{5}-adversarial loss with feature matching~\cite{Goodfellow_nips2014,Radford_iclr2016,salimans_nips2016}, \nth{6}-the combined loss). \nth{7}-appearance flow network (AFN)~\cite{Zhou_eccv2016}. \textbf{\nth{8}-ours(TVSN)}.}
\label{fig:front_ex}
\end{figure*}

We consider the problem of novel 3D view synthesis---given a single view of an object in an arbitrary pose, the goal is to synthesize an image of the object after a specified transformation of viewpoint. It has a variety of practical applications in computer vision, graphics, and robotics. As an image-based rendering technique~\cite{kang_2000}, it allows placing a virtual object on a background with a desired pose or manipulating virtual objects in the scene~\cite{OM3D2014}. Also, multiple generated 2D views form an efficient representation for 3D reconstruction~\cite{tatarchenko_eccv2016}. In robotics, synthesized novel views give the robot a  better understanding of unseen parts of the object through 3D reconstruction, which will be helpful for grasp planning~\cite{Varley_arxiv2016}. 

This problem is generally challenging due to unspecified input viewing angle and the ambiguities of 3D shape observed in only a single view.  In particular inferring the appearances of unobserved parts of the object that are not visible in the input view is necessary for novel view synthesis.  Our approach attacks all of these challenges, but our contributions focus on the later aspect, dealing with disoccluded appearance in novel views and outputting highly-detailed synthetic images.

%We consider the problem of novel 3D view synthesis---given a single view of an object in an arbitrary pose, the goal is the synthesize an image of the object after a specified transformation of viewpoint. This problem is generally challenging due to unspecified input viewing angle and the ambiguities of 3D shape observed in only a single view.  In particular inferring the appearances of unobserved parts of the object that are not visible in the input view is necessary for novel view synthesis.  Our approach attacks all of these challenges, but our contributions focus on the later aspect, dealing with disoccluded appearance in novel views and outputting highly-detailed synthetic images.

Given the eventual approach we will take, using a carefully constructed deep network, we can consider related work on dense prediction with encoder-decoder methods to see what makes the structure of the novel 3D view synthesis problem different.  In particular, there is a lack of pixel-to-pixel correspondences between the input and output view. This, combined with large chunks of missing data due to occlusion, makes novel view synthesis fundamentally different than other dense prediction or generation tasks that have shown promising results with deep networks~\cite{noh_iccv2015, dosovitskiy2015flownet, johnson_eccv2016}. Although the input and desired output views may have similar low-level image statistics, enforcing such constraints directly is difficult.  For example, skip or residual connections, are not immediately applicable as the input and output have significantly different global shapes. Hence, previous 3D novel view synthesis approaches~\cite{yang_nips2015,tatarchenko_eccv2016} have not been able to match the visual quality of geometry-based methods that exploit strong correspondence.

The geometry-based methods are an alternative to pure generation, and have been demonstrated in~\cite{hoiem_togs2005,OM3D2014,rematas2016novel}. Such approaches estimate the underlying 3D structure of the object and apply geometric transformation to pixels in the input (e.g. performing depth-estimation followed by 3D transformation of each pixel~\cite{garg_eccv2016}).  When successful, geometric transformation approaches can very accurately transfer original colors, textures, and local features to corresponding new locations in the target view. However, such approaches are fundamentally unable to hallucinate where new parts are revealed due to disocclusion. Furthermore, even for the visible geometry precisely estimating the 3D shape or equivalently the precise pixel-to-pixel correspondence between input and synthesized view is still challenging and failures can result in distorted output images.

In order to bring some of the power of explicit correspondence to deep-learning-based generation of novel views, the recent appearance flow network (AFN)~\cite{Zhou_eccv2016} trains a convolutional encoder-decoder to learn how to move pixels without requiring explicit access to the underlying 3D geometry. Our work goes further in order to integrate more explicit reasoning about 3D transformation, hallucinate missing sections, and clean-up the final generated image producing significant improvements of realism, accuracy, and detail for synthesized views.

To achieve this we present \emph{a holistic approach to novel view synthesis by grounding the generation process on viewpoint transformation}. Our approach first predicts the transformation of existing pixels from the input view to the view to be synthesized, as well as a visibility map, exploiting the learned view dependency. We use the transformation result matted with the predicted visibility map to condition the generation process. The image generator not only hallucinates the missing parts but also refines regions that suffer from distortion or unrealistic details due to the imperfect transformation prediction. This holistic pipeline alleviates some difficulties in novel view synthesis by explicitly using transformation for the parts where there are strong cues.  % (i.e. parts observed both in the input and the target view).
%This holistic pipeline alleviates difficulties of the novel view synthesis problem by explicitly using transformation for the parts where there are strong cues (i.e. parts observed both in the input and the target view), and using generation for parts where hallucination is necessary. 

We propose an architecture composed of two consecutive convolutional encoder-decoder networks. First, we introduce a disocclusion aware appearance flow network (DOAFN) to predict the visibility map and the intermediate transformation result. Our second encoder-decoder network is an image completion network which takes the matted transformation as an input and completes and refines the novel view with a combined adversarial and feature-reconstruction loss. A wide range of experiments on synthetic and real images show that the proposed technique achieves significant improvement compared to existing methods.
Our main contributions are:
\begin{itemize}
\item We propose a holistic image generation pipeline that explicitly predicts how pixels from the input will be transformed and {\em where there is disocclusion} in the output that needs to be filled, converting the remaining synthesis problem into one of image completion and repair.
\item We design a disocclusion aware appearance flow network that relocates existing pixels in the input view along with predicting a visibility map.  
\item We show that using loss networks with a term considering how well recognition-style features are reconstructed, combined with $L_1$ loss on pixel values during training, improves synthesized image quality and detail.
\end{itemize}

\begin{figure*}[t]
\begin{center}
\includegraphics[width=\linewidth]{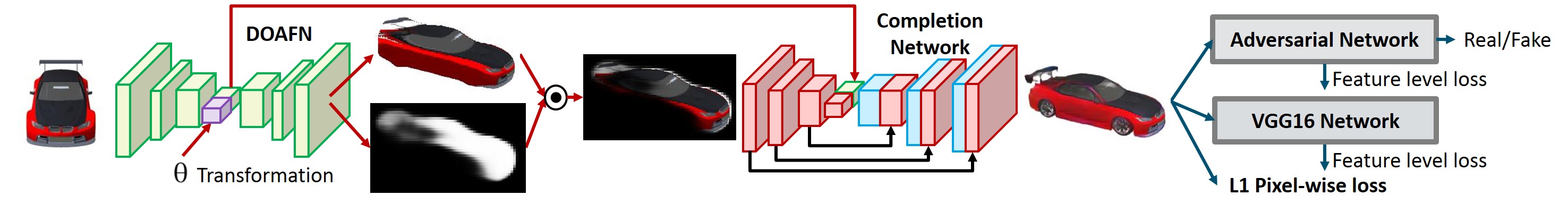}
\end{center}
\caption{Transformation-grounded view synthesis network(TVSN). Given an input image and a target transformation (\ref{sec:doafn}), our disocclusion-aware appearance flow network (DOAFN) transforms the input view by relocating pixels that are visible both in the input and target view. The image completion network, then, performs hallucination and refinement on this intermediate result(\ref{sec:comp}). For training, the final output is also fed into two different loss networks in order to measure similarity against ground truth target view (\ref{sec:lossnet}). }
\label{fig:pipeline}
\end{figure*}

\vspace{-1mm}
\vspace{-1mm}
\section{Related Work}
%\textbf{Image generation via iterative optimization}. 
%\textbf{Generative Adversarial Networks}.
%\textbf{Geometric image transformation}.
%\textbf{Image completion}.
%\textbf{Image generation with deep CNNs}.

%\paragraph{Learning image transformation.}
%Tranformating auto-encoders~\cite{hinton2011transforming}. Transformer networks~\cite{jaderberg_NIPS2015}. Dynamic filters networks~\cite{de2016dynamic}. Single-view depth estimation~\cite{garg_eccv2016}. Appearance flow networks~\cite{Zhou_eccv2016}. DeepStereo~\cite{flynn2015deepstereo}. Deep3D~\cite{xie2016deep3d}.

\vspace{-1mm}
\paragraph{Geometry-based view synthesis.}
A large body of work benefits from implicit or explicit geometric reasoning to address the novel view synthesis problem. When multiple images are available, multi-view stereo algorithms~\cite{furukawa_mvs} are applicable to explicitly reconstruct the 3D scene which can then be utilized to synthesize novel views. An alternative approach recently proposed by Flynn et al.~\cite{flynn2015deepstereo} uses deep networks to learn to directly interpolate between neighboring views. Ji et al.~\cite{ji_cvpr2017} propose to rectify the two view images first with estimated homography by deep networks, and then synthesize middle view images with another deep networks. In case of single input view, Garg et al.~\cite{garg_eccv2016} propose to first predict a depth map and then synthesize the novel view by transforming each reconstructed 3D point in the depth map. However, all these approaches only utilize the information available in the input views and thus fail in case of disocclusion. Our method, on the other hand, not only takes advantage of implicit geometry estimation but also infers the parts of disocclusion.

Another line of geometry-based methods utilize large internet collections of 3D models which are shown to cover wide variety for certain real world object categories~\cite{OM3D2014,rematas2016novel}. Given an input image, these methods first identify the most similar 3D model in a database and fit to the image either by 3D pose estimation~\cite{rematas2016novel} or manual interactive annotation~\cite{OM3D2014}. The 3D information is then utilized to synthesize novel views. While such methods generate high quality results when sufficiently similar 3D models exist, they are often limited by the variation of 3D models found in the database. In contrast, our approach utilizes 3D models only for training generation networks that directly synthesize novel views from an image.

\vspace{-1mm}
\paragraph{Image generation networks.}
One of the first convolutional networks capable of generating realistic images of objects is proposed in~\cite{Dosovitskiy_cvpr2015}, but the network requires explicitly factored representations of object type, viewpoint and color, and thus is not able to generalize to unseen objects. 
The problem of generating novel views of an object from a single image is addressed in~\cite{yang_nips2015,kulkarni_nips2015,tatarchenko_eccv2016} using deep convolutional encoder-decoder networks. Due to the challenges of disentangling the factors from single-view and the use of globally smooth pixel-wise similarity measures (e.g. $L_1$ or $L_2$ norm), the generation results tend to be blurry and low in resolution. 

An alternative to learning disentangled or invariant factors is the use of equivariant representations, i.e. transformations of input data which facilitate downstream decision making. Transforming auto-encoders are coined by Hinton et al.~\cite{hinton2011transforming} to learn both 2D and 3D transformations of simple objects. Spatial transformer networks~\cite{jaderberg_NIPS2015} further introduce differentiable image sampling techniques to enable in-network parameter-free transformations. In the 3D case, flow fields are learned to transform input 3D mesh to the target shape~\cite{yumer2016learning} or input view to the desired output view~\cite{Zhou_eccv2016}. However, direct transformations are clearly upper-bounded by the input itself. To generate novel 3D views, our work grounds a generation network on the learned transformations to hallucinate disoccluded pixels.

Recently, a number of image generation methods introduce the idea of using pre-trained deep networks as loss function, referred as perceptual loss, to measure the feature similarities from multiple semantic levels~\cite{johnson_eccv2016,larsen_icml2016,ulyanov_icml2016,lamb2016discriminative}. The generation results from these works well preserve the object structure but are often accompanied with artifacts such as aliasing. At the same time, generative adversarial networks~\cite{Goodfellow_nips2014,Radford_iclr2016}, introduce a discriminator network, which is adversarially trained with the generator network to tell apart the generated images from the real ones. The discriminator encapsulates natural image statistics of all orders in a real/fake label, but its min-max training often leads to local minimum, and thus local distortions or painting-stroke effects are commonly observed in their generated images. Our work uses a combined loss function that takes advantages of both the structure-preserving property of perceptual loss and the rich textures of adversarial loss (See Fig.~\ref{fig:teaser}).

Deep networks have also been explored for image completion purposes. Examples of proposed methods include image in-painting with deep networks~\cite{pathak_cvpr2016} and sequential parts-by-parts generation for image completion~\cite{kwak_arxiv2016}. Such methods assume the given partial input is correct and focus only on completion. In our case, however, we do not have access to a perfect intermediate result. Instead, we rely on the generation network both to hallucinate missing regions and also refine any distortions that occur due to inaccurate per-pixel transformation prediction.

\begin{figure*}[t]
\begin{center}
\includegraphics[width=\linewidth]{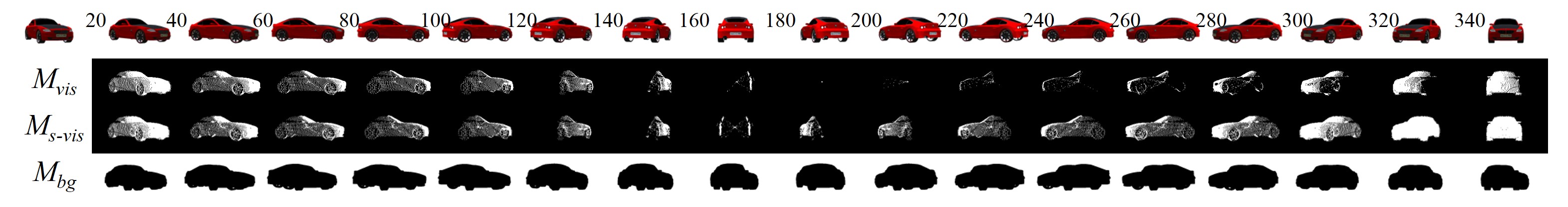}
\end{center}
\caption{Visibility maps of different rotational degrees: the first column in the first row is an input image. Rest of columns show output images and corresponding masks given transformation ranging from 20 to 340 rotational degrees with 20 degree intervals. The second, third and fourth rows show visibility maps $M_{vis}$, symmetry-aware visibility maps $M_{s-vis}$, and background masks $M_{bg}$ respectively. The input image is in the pose of 0 elevation and 20 azimuth. The visibility maps of the rotational degree from 160 to 340 show the main difference between $M_{vis}$ and $M_{s-vis}$. For example, we assume the opposite side of the car visible with $M_{s-vis}$ even if those parts were not seen in the input image.}
\label{fig:masks}
\vspace{-2mm}
\end{figure*}

\section{Transformation-Grounded View Synthesis}
%
%Our goal is to synthesize novel 3D views of an object from a single image simulating a virtual camera moving around the object. 
Novel view synthesis could be seen as a combination of the following three scenarios: 1) pixels in the input view that remain visible in the target view are moved to their corresponding positions; 2) remaining pixels in the input view disappear due to occlusions; and 3) previously unseen pixels are revealed or disoccluded in the target view. We replicate this process via a neural network as shown in Figure~\ref{fig:pipeline}. Specifically, we propose a disocclusion-aware appearance flow network (\ref{sec:doafn}) to transform the pixels of the input view that remain visible. A subsequent generative completion network (\ref{sec:comp}) then hallucinates the unseen pixels of the target view given these transformed pixels.

\subsection{Disocclusion-aware Appearance Flow Network}
\label{sec:doafn}
Recently proposed appearance flow network (AFN)~\cite{Zhou_eccv2016} learns how to move pixels from an input to a target view. The key component of the AFN is a differentiable image sampling layer introduced in \cite{jaderberg_NIPS2015}. Precisely, the network first predicts a dense flow field that maps the pixels in the target view, $I_t$, to the source image, $I_s$. Then, sampling kernels are applied to get the pixel value for each spatial location in $I_t$. Using a bilinear sampling kernel, the output pixel value at spatial location $I_t^{i,j}$ equals to:
%\begin{equation}
%I_t^{(i,j)}\!\!=\!\!\sum_{(h,w) \in N} I_s^{(h,w)} %\max(0,1-|F_y^{(i,j)}-h|) \max(0,1-|F_x^{(i,j)}-w|),
%\end{equation}
\begin{equation}
\sum_{(h,w) \in N} I_s^{h,w} \max(0,1-|F_y^{i,j}-h|) \max(0,1-|F_x^{i,j}-w|),
\end{equation}
where $F$ is the flow predicted by the deep convolutional encoder-decoder network (see the first half of Figure~\ref{fig:pipeline}). $F_x^{i,j}$ and $F_y^{i,j}$ indicate the $x$ and $y$ coordinates of one target location. $N$ denotes the 4-pixel neighborhood of $(F_y^{i,j},F_x^{i,j})$. 
%Using L1 norm as a loss function, $L = ||I_{gt} - I_{t}||_1$, the subgradients w.r.t predicted coordinates $\frac{\partial L}{\partial F}$ are available so that standard backpropagation procedure can be adopted to compute the gradients w.r.t network parameters.
%$\frac{\partial L}{\partial \theta}$.

The key difference between our disocclusion aware appearance flow network (DOAFN) and the AFN is the prediction of an additional visibility map which encodes the parts that need to be removed due to occlusion. The original AFN synthesizes the entire target view, including the disoccluded parts, with pixels of the input view, e.g. \nth{1} row of AFN results in Figure~\ref{fig:front_ex}. However, such disoccluded parts might get filled with wrong content, resulting in implausible results, especially for cases where a large portion of the output view is not seen in the input view. Such imperfect results would provide misleading information to a successive image generation network. Motivated by this observation, we propose to predict a visibility map that masks such problematic regions in the transformed image:
\begin{equation}
I_{doafn} = I_{afn} \odot M_{vis},
\end{equation}
where $M_{vis} \in [0,1]^{H \times W}$. 
To achieve this, we define the ground truth visibility maps according to the 3D object geometry as described next.
%We choose to use a differentiable soft mask so that we can apply the backpropagation algorithm for training two consecutive networks jointly. 

\vspace{-2mm}
\paragraph{Visibility map.}
Let $M_{vis} \in \mathbb{R}^{H \times W}$ be the visibility map for the target view, given source image $I_s$ and desired transformation parameter $\theta$. The mapping value for a pixel in the target view corresponding to a spatial location $(i,j)$ in $I_s$ is defined as follows: 
\begin{equation}
M_{vis}^{(PR(\theta)\mathbf{x}_s^{(i,j)})^h,(PR(\theta)\mathbf{x}_s^{(i,j)})^w} = \begin{cases} 
1 & \mathbf{c}^\top R(\theta)\mathbf{n}_s^{(i,j)} > 0\\
0 & \text{otherwise}
\end{cases}
\label{eq:dis}
\end{equation}
$\mathbf{x}_s^{(i,j)} \in \mathbb{R}^{4}$ is the 3D object coordinates and $\mathbf{n}_s^{(i,j)} \in \mathbb{R}^{4}$ is the surface normal corresponding to location $(i,j)$ in $I_s$, both represented in homogeneous coordinates. Since we use synthetic renderings of 3D CAD models, we have access to ground truth object coordinates and surface normals. $R(\theta) \in \mathbb{R}^{3 \times 4}$ is the rotation matrix given the transformation parameter $\theta$ and $P \in \mathbb{R}^{3 \times 3}$ is the perspective projection matrix. The superscripts $h$ and $w$ denote the target image coordinates in $y$ and $x$ axis respectively after perspective projection. $\mathbf{c} \in \mathbb{R}^{3}$ is the 3D camera center. In order to compute the target image coordinates for each pixel in $I_s$, we first obtain the 3D object coordinates corresponding to this pixel and then apply the desired 3D transformation and perspective projection. The mapping value of the target image coordinate is 1 if and only if the dot product between the viewing vector and surface normal is positive, i.e. the corresponding 3D point is pointing towards the camera.

\vspace{-2mm}
\paragraph{Symmetry-aware visibility map.}
Many common object categories exhibit reflectional symmetry, e.g. cars, chairs, tables etc. 
%AFN implicitly exploits this characteristic to successfully generate disoccluded regions utilizing their visible symmetric counterparts. In fact, for certain desired transformations AFN learns to perform simple mirroring to generate the perfect target image. In order to fully take advantage of this strength of AFN, we propose to use a symmetry-aware visibility map.
AFN implicitly exploits this characteristic to ease the synthesis of large viewpoint changes. To fully take advantage of symmetry in our DOAFN, we propose to use a symmetry-aware visibility map.
Assuming that objects are symmetric with respect to the $xy$-plane, a symmetry-aware visibility map $M_{sym}$ is computed by applying Equation~\ref{eq:dis} to the z-flipped object coordinates and surface normals. The final mapping for a pixel in the target view corresponding to spatial location $(i,j)$ is then defined as:
\begin{equation}
M_{s-vis}^{i,j} = \mathbbm{1}{\left[ M_{sym}^{i,j} + M_{vis}^{i,j} > 0 \right]}
\end{equation}

\paragraph{Background mask.}
%In case of images depicting an object on a natural background, 
%transforming the object also reveals part of the background not visible in the input view.
% (e.g., rotating a side view of a car to be frontal). Therefore, we define the background to be only those regions visible both in input and output view 
%we only transforms the foreground objects and backgrounds remain unchanged. 
%This is practically useful scenario, such as 3D photo editing application. 
Explicit decoupling of the foreground object is necessary to deal with real images with natural background. In addition to parts of the object being disoccluded in the target view, different views of the object occlude different portions of the background posing additional challenges. For example, transforming a side view to be frontal exposes parts of the background occluded by the two ends of the car. In our approach, we define the \emph{foreground} as the region that covers pixels of the object in both input view and output view. The rest of the image belongs to the \emph{background} and should remain unchanged in both views. We thus introduce a unified background mask, 
\begin{equation}
M_{bg}^{i,j} = \mathbbm{1}{\left[ B_s^{i,j} + B_t^{i,j} > 0 \right]},
\end{equation}
where $B_s$ and $B_t$ are the background masks of the source and target images respectively. Ground truth background masks are easily obtained from 3D models. Examples of background masks are presented in Figure~\ref{fig:masks}.
When integrated with the (symmetry-aware) visibility map, the final output of DOAFN becomes:
\begin{equation}
I_{doafn} = I_{s} \odot M_{bg} + I_{afn} \odot M_{s-vis}
\end{equation}

\vspace{-2mm}
\subsection{View Completion Network}
\label{sec:comp}
Traditional image completion or hole filling methods often exploit local image information \cite{efros_siggraph2001,barnes_siggraph2009,wexler_tpami2007} and have shown impressive results for filling small holes or texture synthesis. In our setting, however, sometimes more than half of the content in the novel view is not visible in the input image, constituting a big challenge for local patch based methods. To address this challenge, we propose another encoder-decoder network, capable of utilizing both local and global context, to complete the transformed view inferred by DOAFN.

%Deep networks have also been explored in image completion domain. Image inpainting with deep networks has shown promising results\cite{pathak_cvpr2016} and parts by parts generation sequentially to complete whole image was suggested \cite{kwak_arxiv2016}. A common theme of their network designs is that the completion network is only allowed to fill the parts that were not filled in the inputs. In other words, the network believes that contents in the inputs are correct. However, this design principle is not directly applicable to our completion network since the input, which is output of DOAFN, is sometimes highly distorted.
%
Our view completion network is composed of an ``hourglass" architecture similar to~\cite{newell_eccv2016}, with a bottleneck-to-bottleneck identity mapping layer from DOAFN to the hourglass (see Figure~\ref{fig:pipeline}). This network has three essential characteristics. First, being conditioned on the high-level features of DOFAN, it can generate content that have consistent attributes with the given input view, especially when large chunk of pixels are dis-occluded. Second, the output of DOAFN is already in the desired viewpoint with important low-level information, such as colors and local textures, preserved under transformation. Thus, it is possible to utilize skip connections to propagate this low-level information from the encoder directly to later layers of the decoder. Third, the view completion network not only hallucinates disoccluded regions but also fixes artifacts such as distortions or unrealistic details. The output quality of DOAFN heavily depends on the input viewpoint and desired transformation, resulting in imperfect flow in certain cases. The encoder-decoder nature of the image generation network is well-suited to fix such cases. Precisely, while the encoder is capable of recognizing undesired parts in the DOAFN output, the decoder refines these parts with realistic content.

\vspace{-2mm}
\paragraph{Loss networks.}
\label{sec:lossnet}
The idea of using deep networks as a loss function for image generation has been proposed in \cite{larsen_icml2016,ulyanov_icml2016,johnson_eccv2016}. Precisely, an image generated by a network is passed as an input to an accompanied network which evaluates the discrepancy (the feature distance) between the generation result and ground truth. We use the \textit{VGG16} network for calculating the feature reconstruction losses from a number of layers, which is referred as \emph{perceptual loss}.
%First, images are generated by deep networks. Generated images then go through the second network, to get the signal of how far generated images from the ground truth images in the loss network point of view. It can be usually done by extracting features of generated images and ground truth images from the loss network and computing the feature reconstruction loss. We used \textit{vgg16} pre-trained network to measure the perceptual similarity.
%
%In our initial experiments, we utilized a loss network with random weights suggested in \cite{he_nips2016,Ustyuzhaninov_arxiv2016}. However, we discovered that 
We tried both a pre-trained loss network and a network with random weights as suggested in~\cite{he_nips2016,Ustyuzhaninov_arxiv2016}. However, we got perceptually poor results with random weights, concluding that the weights of the loss network indeed matter. 

On the other hand, adversarial training~\cite{Goodfellow_nips2014} has been phenomenally successful for training the loss network at the same time of training the image generation network. We experimented with a similar adversarial loss network as in~\cite{Radford_iclr2016} while adopting the idea of feature matching presented in~\cite{salimans_nips2016} to make the training process more stable.

We realized that the characteristics of generated images with these two kinds of loss networks, perceptual and adversarial, are complementary. Thus, we combined them together with the standard image reconstruction loss ($L_1$) to maximize performance. Finally, we added total variation regularization term ~\cite{johnson_eccv2016}, which was useful to refine the image:
\begin{multline}
-\log D(G(I_s)) + \alpha L_{2}(F_{D}(G(I_s)), F_{D}(I_t)) ) + \\
\beta L_{2}(F_{vgg}(G(I_s)), F_{vgg}(I_t)) + \gamma L_1(I_s,I_t) + \lambda L_{TV}(G(I_s))
\end{multline}
$I_s$, $G(I_s)$ and $I_t$ is the input, generated output and corresponding target image, respectively. $\log(D)$ is log likelihood of generated image $G(I_s)$ being a real image, estimated by adversarially trained loss network, called discriminator $D$. In practice, minimizing $-\log D(G(I_s))$ has shown better gradient behaviour than minimizing $\log D(1 - G(I_s))$.

$F_D$ and $F_{vgg}$ are the features extracted from the discriminator and \textit{VGG16} loss networks respectively. We found that concatenated features from the first to the third convolutional layers are the most effective. $L_1$ and $L_2$ are $\ell_1$ and $\ell_2$ norms of two same size inputs divided by the size of the inputs. In sum, both generated images $G(I_s)$ and ground truth image $I_t$ are fed into $D$ and \textit{VGG16} loss networks, and we extract the features, and compute averaged euclidean distance between these two. 

The discriminator $D$ is simultaneously trained along with $G$ via alternative optimization scheme proposed in~\cite{Goodfellow_nips2014}. The loss function for the discriminator is 
\begin{equation}
-\log D(I_s) - \log (1-D(G(I_s)))
\end{equation}
We empirically found that $\alpha=100$, $\beta=0.001$, $\gamma=1$, and $\lambda=0.0001$ are good hyper-parameters and fixed them for the entire experiments.

\begin{figure}[t]
\begin{center}
\includegraphics[width=\linewidth]{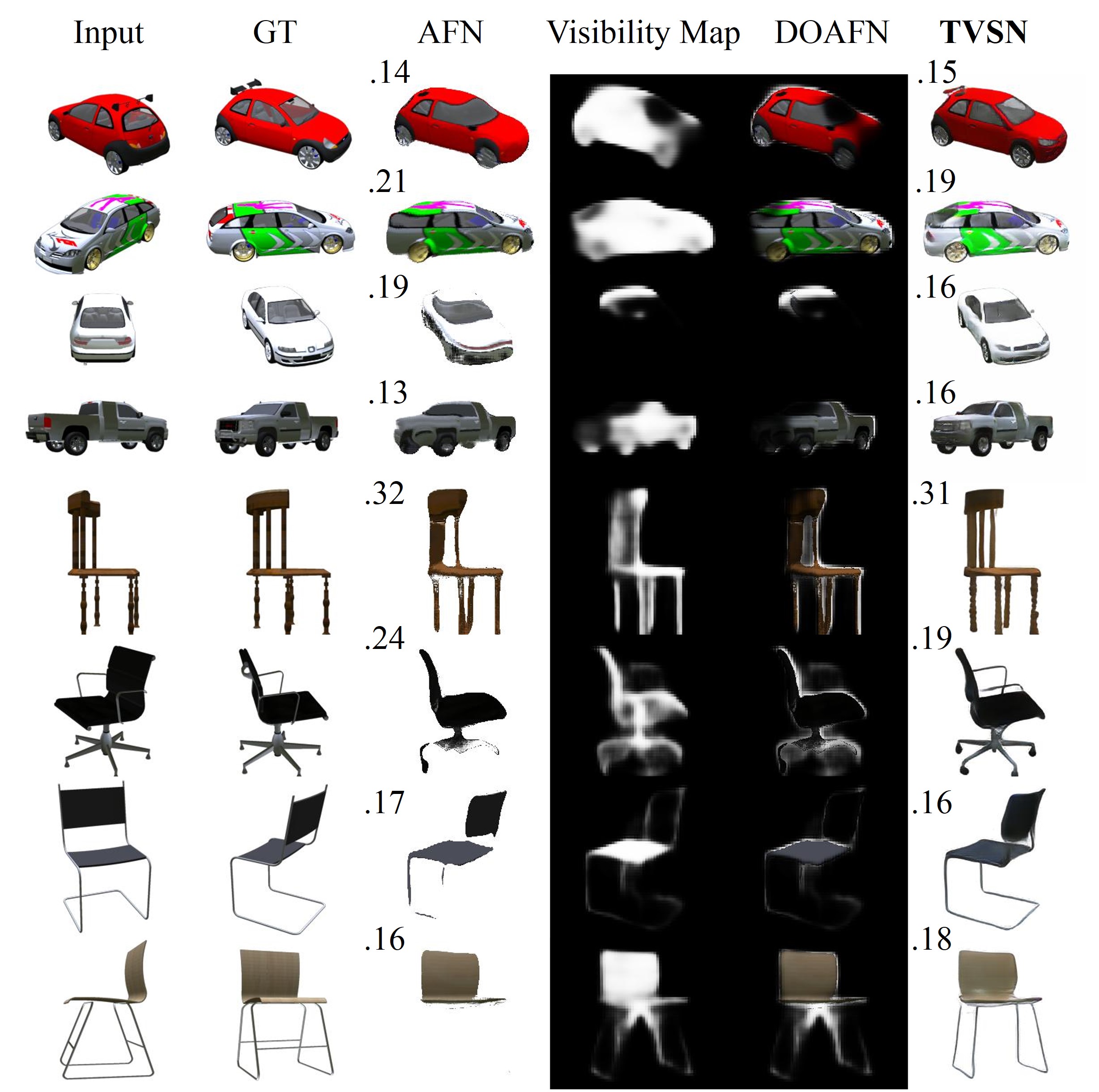}
\end{center}
\caption{Results on synthetic data from ShapeNet. We show the input, ground truth output (GT), results for AFN and our method (TVSN) along with the $L_1$ error. We also provide the intermediate output (visibility map and output of DOAFN).}
\label{fig:tvsn_results}
\vspace{-2mm}
\vspace{-2mm}
\end{figure}

\section{Experiments}
\subsection{Training Setup}
We use rendered images of 3D models from ShapeNet~\cite{shapenet2015} both for training and testing. We use the entire \emph{car} category ($7497$ models) and a subset of the \emph{chair} category ($698$ models) with sufficient texture. For each model, we render images from a total of $54$ viewpoints corresponding to $3$ different elevations ($0$, $10$, and $20$) and $18$ azimuth angles (sampled in the range $[0,340]$ with $20$-degree increments). The desired transformation is encoded as a 17-D one-hot vector corresponding to one of the rotation angles between input and output views in the range $[20,340]$. Note that we did not encode 0 degree as it is the identical mapping. For each category, 80\% of 3D models are used for training, which leaves over $5$ million training pairs (input view-desired transformation) for the car category and $0.5$ million for the chair category. We randomly sample input viewpoints, desired transformations from the rest 20\% of 3D models to generate a total of $20,000$ testing instances for each category. Both input and output images are of size $256\!\times\!256\!\times\!3$. 
%
%We used the 3D models from the ShapeNet~\cite{shapenet2015} and the images were rendered in multiple view points (azimuths ranging from 0 to 340 at steps of 20 degree, and 0, 10 and 20 degrees of elevations). For transformation, we used 17 discrete azimuth rotations, ranging from 20 to 340 with 20 steps. We encode the transformation as a 17-D one hot vector. We used entire car category(7497 instances), and for chair category, selected 698 instances that have meaningful textures\cite{Zhou_eccv2016}. We randomly picked 80\% of datasets as training set and rest of them are test set. For all experiments our network takes 256 x 256 size images as inputs and produces same size of outputs. When training, we randomly sample the 3D models, input view points, and rotational degrees. For testing, we used random 20,000 samples. For images with backgrounds, we picked up the random images from SUN397 dataset\cite{Su_iccv2015} and randomly cropped the images with the size of 256 x 256, which is the same as the image size. We implemented all baselines used in the paper by ourselves.

We first train DOAFN, and then the view completion network while DOAFN is fixed. After the completion network fully converges, we fine-tune both networks end-to-end. However, this last fine-tuning stage does not show notable improvements. We use mini-batches of size $25$ and $15$ for DOAFN and the completion network respectively. The learning rate is initialized as $1^{-4}$ and is reduced to $1^{-5}$ after $100,000$ iterations. For adversarial training, we adjust the update schedule (two iterations for generator and one iteration for discriminator in one cycle) to balance the discriminator and the generator.

\begin{table}[]
\centering
\caption{We compare our method (\emph{TVSN(DOAFN)}) to several baselines: (i) a single-stage encoder-decoder network trained with different loss functions: $L_1$ (\emph{$L_1$}), feature reconstruction loss using VGG16 (\emph{VGG16}), adversarial (\emph{Adv}), and combination of the latter two (\emph{VGG16+Adv}), (ii) a variant of our approach that does not use a visibility map (\emph{TVSN(AFN)}).}
\label{tabel:quant}
\begin{tabular}{l|cc|cc}
                  & \multicolumn{2}{c}{car} & \multicolumn{2}{|c}{chair} \\ \hline
                  & $L_1$     & SSIM   & $L_1$      & SSIM   \\ \hline
$L_1$\cite{tatarchenko_eccv2016}                & .168  & .884  & .248   & \textbf{.895} \\
VGG               & .228  & .870  & .283   & \textbf{.895} \\
Adv               & .208  & .865  & .241   & .885 \\
VGG+Adv           & .194  & .872  & .242   & .888 \\
AFN\cite{Zhou_eccv2016}               & .146  & .906  & .240   & .891 \\
TVSN(AFN)         & \textbf{.132} & \textbf{.910} & \textbf{.229} & \textbf{.895}  \\
TVSN(DOAFN)       & \textbf{.133} & \textbf{.910} & \textbf{.230} & \textbf{.894}  \\
\end{tabular}
\vspace{-2mm}
\vspace{-2mm}
\end{table}

\subsection{Results}
We discuss our main findings in the rest of this section and refer the reader to the supplementary material for more results.
We utilize the standard $L_1$ mean pixel-wise error and the structural similarity index measure (SSIM)~\cite{Wang04imagequality,mathieu_iclr2016} for evaluation. When computing the $L_1$ error, we normalize the pixel values resulting in errors in the range $[0,1]$, lower numbers corresponding to better results. SSIM is in the range $[-1,1]$ where higher values indicate more structural similarity.

\label{sec:tvsn_results_afn}
\vspace{-2mm}
\begin{figure}[thbp]
\begin{center}
\includegraphics[width=\linewidth]{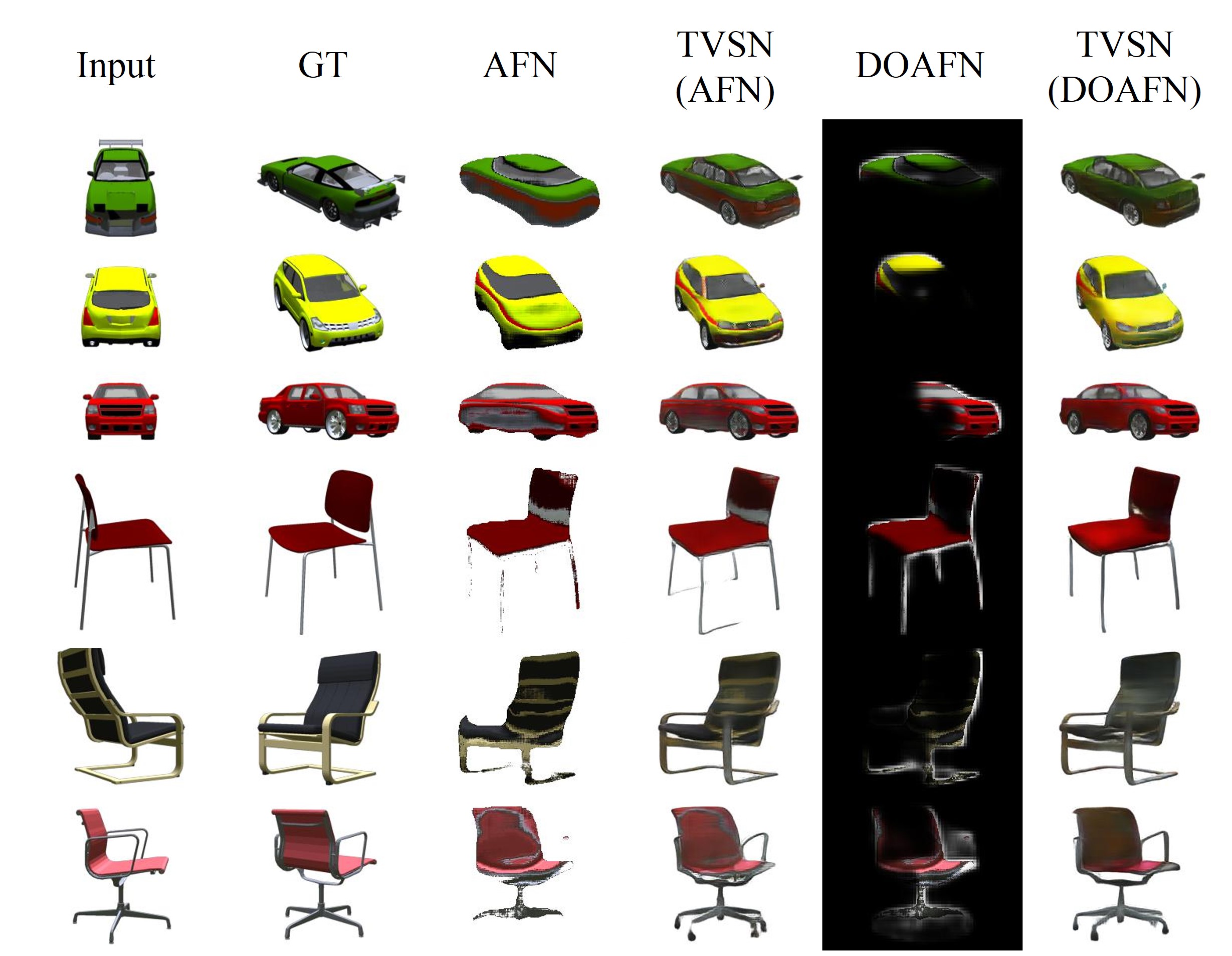}
\end{center}
\caption{When a visibility map is not utilized (TVSN(AFN)), severe artifacts observed in the AFN output get integrated into the final results. By masking out such artifacts, our method (TVSN(DOAFN)) relies purely on the view completion network to generate plausible results.}
\label{fig:tvsn_results_afn}
\vspace{-3mm}
\end{figure}

\vspace{-2mm}
\paragraph{Comparisons.} We first evaluate our approach on synthetic data and compare to AFN. Figure~\ref{fig:tvsn_results} shows qualitative results.\footnote{The results from the original AFN~\cite{Zhou_eccv2016} paper are not directly comparable due to the different image size. In addition, since the complete source code was not available at the time of paper submission, we re-implemented this method by consulting the authors.}
We note that while our method completes the disoccluded parts consistently with the input view, AFN generates unrealistic content (front and rear parts of the cars in the 1st and 2nd rows). Our method also corrects geometric distortions induced by AFN (3rd and 4th rows) and better captures the lighting (2nd row). For the chair category, AFN often fails to generate thin structures such as legs due to the small number of pixels in these regions contributing to the loss function. On the other hand, both perceptual and adversarial loss help to complete the missing legs as they contribute significantly to the perception of the overall shape.

\vspace{2mm}
In order to evaluate the importance of the visibility map, we compare against a variant of our approach which directly provides the output of AFN to the view completion network without masking. (For clarity, we will refer to our method as \emph{TVSN(DOAFN)} and to this baseline as \emph{TVSN(AFN)}.) Furthermore, we also implement a single-stage convolutional encoder-decoder network as proposed in~\cite{tatarchenko_eccv2016} and train it with various loss functions: $L_1$ loss ($L_1$), feature reconstruction loss using VGG16 (VGG16), adversarial loss (Adv), and combination of the latter two (VGG16+Adv).
%Even for the cases where outputs of DOAFN are severely distorted or warped, our TVSN completed and refined the images surprisingly well. Based on this observation, we implemented another baseline that discards the masking stage with the visibility map prediction. Therefore, entire output pixels provided by AFN are fed into the image completion network, which is called TVSN(AFN) as opposed to TVSN(DOAFN).
We provide quantitative and visual results in Table~\ref{tabel:quant} and Figure~\ref{fig:front_ex} respectively. We note that, although commonly used, $L_1$ and SSIM metrics are not fully correlated with human perception. While our method is clearly better than the $L_1$ baseline~\cite{tatarchenko_eccv2016}, both methods get comparable SSIM scores. 

We observe that both TVSN(AFN) and TVSN(DOAFN) perform similarly with respect to $L_1$ and SSIM metrics demonstrating that the view completion network in general successfully refines the output of AFN. However, in certain cases severe artifacts observed in the AFN output, especially in the disoccluded parts, get smoothly integrated in the completion results as shown in Figure~\ref{fig:tvsn_results_afn}. In contrast, the visibility map masks out those artifacts and thus TVSN(DOAFN) relies completely on the view completion network to hallucinate these parts in a realistic and consistent manner.

\begin{figure}[t]
\vspace{-2mm}
\begin{center}
\includegraphics[width=\linewidth]{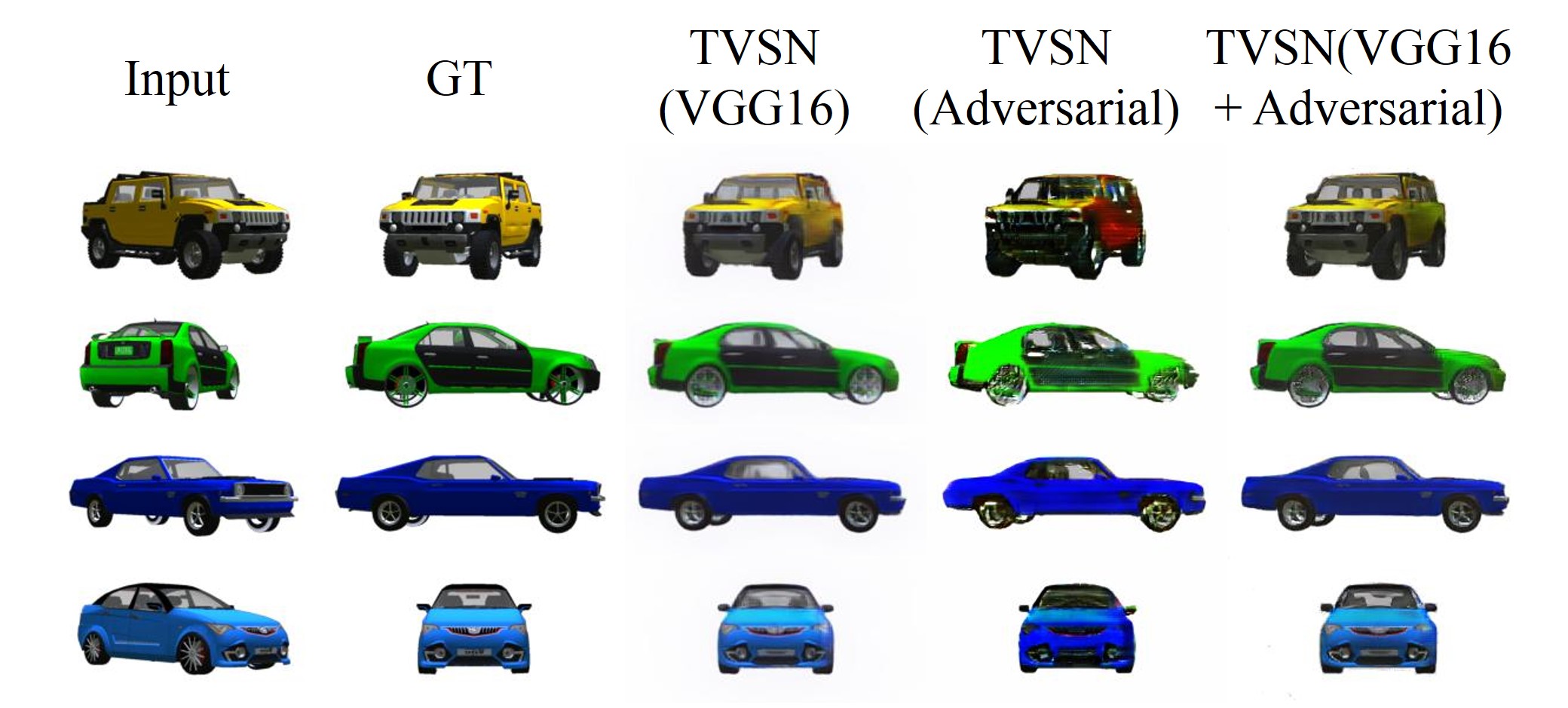}
\end{center}
\caption{We evaluate the effect of utilizing VGG16, (TVSN(VGG16)), and adversarial loss, (TVSN(Adversarial)), only as opposed to our method, (TVSN(VGG16+Adversarial)), which uses a combination of both.}
\label{fig:tvsn_results_losses}
\vspace{-3mm}
\end{figure}

\begin{figure*}[t]
\vspace{-2mm}
\begin{center}
\includegraphics[width=\linewidth]{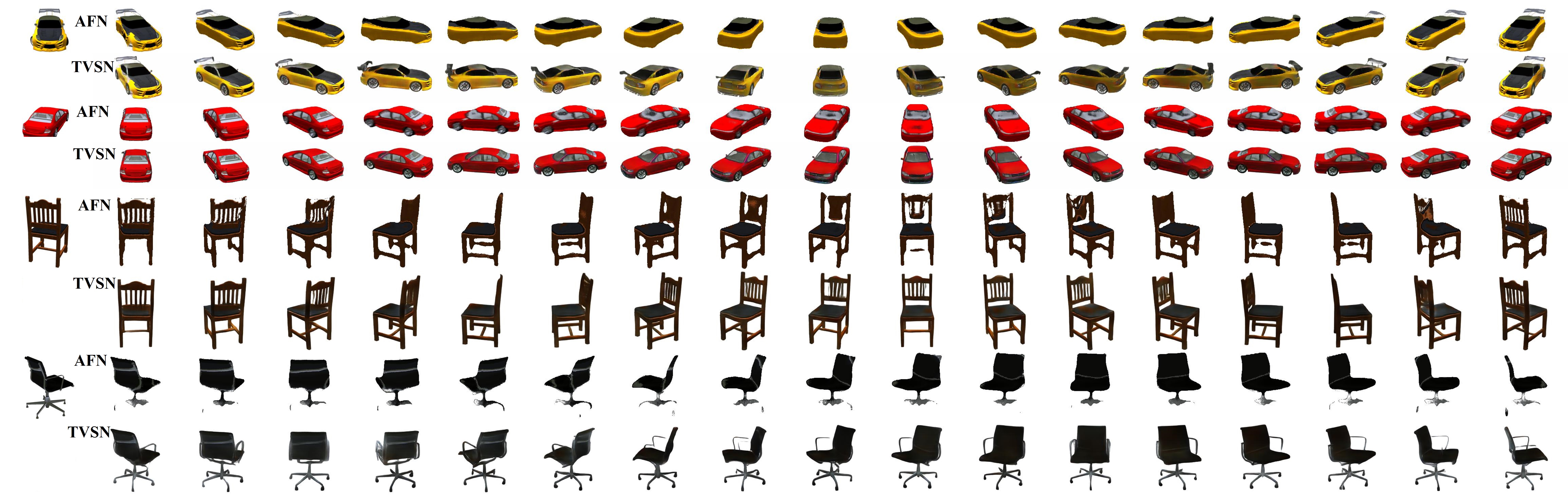}
\end{center}
\caption{Results of 360 degree rotations}
\label{fig:tvsn_results_seq}
\vspace{-2mm}
\end{figure*}

\vspace{-3mm}
\paragraph{Evaluation of the Loss Networks.} We train our network utilizing the feature reconstruction loss of VGG16 and the adversarial loss. We evaluate the effect of each loss by training our network with each of them only and provide visual results in Figure~\ref{fig:tvsn_results_losses}. It is well-known that the adversarial loss is effective in generating realistic and sharp images as opposed to standard pixel-wise loss functions. However, some artifacts such as colors and details inconsistent with the input view are still observed. For the VGG16 loss, we experimented with different feature choices and empirically found that the combination of the features from the first three layers with total variation regularization is the most effective. Although the VGG16 perceptual loss is capable of generating high quality images for low-level tasks such as super-resolution, it has not yet been fully explored for pure image generation tasks as required for hallucinating disoccluded parts. Thus, this loss still suffers from the blurry output problem whereas combination of both VGG16 and adversarial losses results in the most effective configuration.

\begin{figure}[t]
\vspace{-2mm}
\begin{center}
\includegraphics[width=\linewidth]{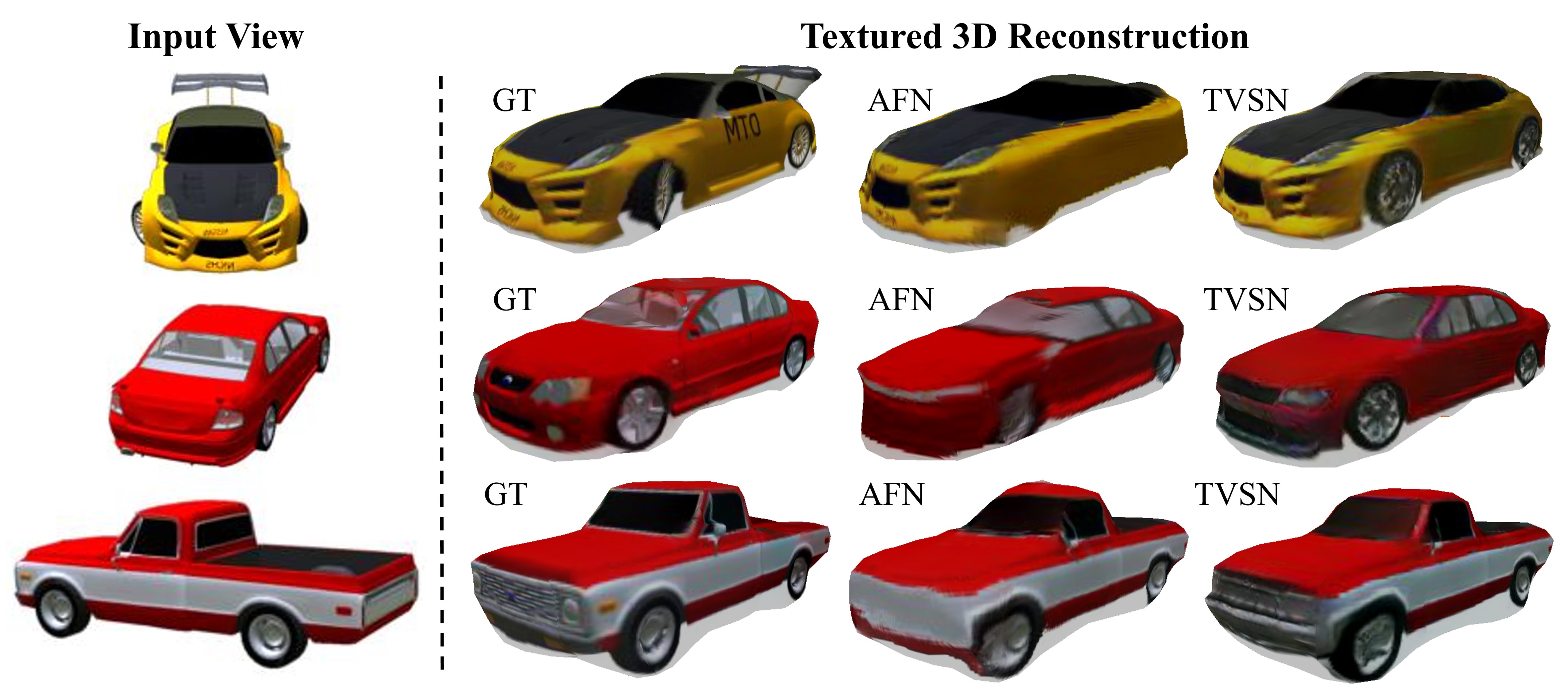}
\end{center}
\caption{We run a multi-view stereo algorithm to generate textured 3D reconstructions from a set of images generated by AFN and our TVSN approach. We provide the reconstructions obtained from ground truth images (GT) for reference.}
\label{fig:3d_reconstruction}
\vspace{-2mm}
\vspace{-2mm}
\end{figure}

\subsection{360 degree rotations and 3D reconstruction}

%\duygu{the following paragraph can be removed if we want to save space, here we are just showing an application of our method, we don't need an explicit comparison}
Inferring 3D geometry of an object from a single image is the holy-grail of computer vision research. Recent approaches using deep networks commonly use a voxelized 3D reconstruction as output~\cite{choy_eccv2016,we_nips2016}. However, computational and spatial complexities of using such voxelized representations in standard encoder-decoder networks significantly limits the output resolution, e.g. $32^3$ or $64^3$. %In addition, It gets worse if we considered color channels for each voxels.

Inspired by~\cite{tatarchenko_eccv2016}, we exploit the capability of our method in generating novel views for reconstruction purposes. Specifically, we generate multiple novel views from the input image to cover a full 360 rotation around the object sampled at $20$-degree intervals. We then run a multi-view reconstruction algorithm~\cite{furukawa_mvs} on these images using the ground truth relative camera poses to obtain a dense point cloud. We use the open source OpenMVS library~\cite{openMVS} to reconstruct a textured mesh from this point cloud. Figure~\ref{fig:tvsn_results_seq} shows multi-view images generated by AFN and our method whereas Figure~\ref{fig:3d_reconstruction} demonstrates examples of reconstructed 3D models from these images. By generating views consistent in terms of geometry and details, our method results in significantly better quality textured meshes.
%Our method preserves more details in the input image, hallucinates appropriately (front views for the car), and refines the images into more plausible ones. Figure \ref{fig:3d_reconstruction} depicted examples of reconstructed 3D models from generated images. TVSN outperforms AFN and this indicates that our method generates not only high quality images but also geometrically consistent ones across multiple views.

%
\subsection{3D Object Rotations in Real Images}
\begin{figure}[t]
\vspace{-2mm}
\begin{center}
\includegraphics[width=\linewidth]{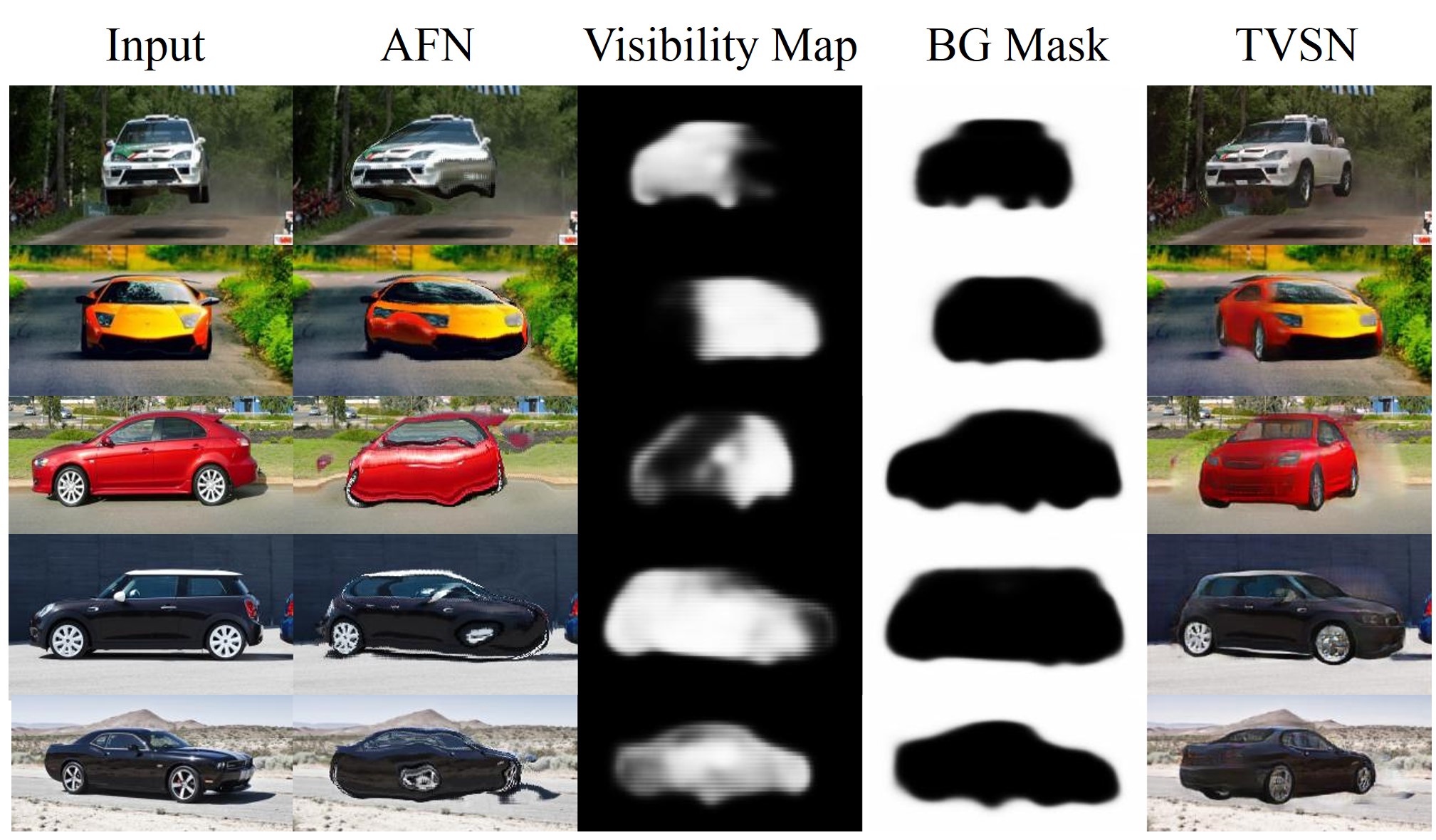}
\end{center}
\caption{We show novel view synthesis results on real internet images along with the predicted visibility map and the background mask.}
\label{fig:tvsn_results_bg}
\vspace{-2mm}
\end{figure}

In order to generalize our approach to handle real images, we generate training data by compositing synthetic renderings with random backgrounds~\cite{Su_iccv2015}. We pick $10,000$ random images from the SUN397 dataset\cite{Su_iccv2015} and randomly crop them to be of size $256\!\times\!256\!\times\!3$. Although this simple approach fails to generate realistic images, e.g. due to inconsistent lighting and viewpoint, it is effective in enabling the network to recognize the contours of the objects in complex background. In Figure~\ref{fig:tvsn_results_bg}, we show several novel view synthesis examples from real images obtained from the internet. 
%We also provide the visibility maps and background masks predicted by DOAFN. We note that we crop the real images to match the size of the network input while keeping the objects roughly at the center similar to our training data.

While our initial experiments show promising results, further investigation is necessary to improve performance. Most importantly, more advanced physically based rendering techniques are required to model complex light interactions in the real world (e.g. reflections from the environment onto the object surface). In addition, it is necessary to sample more viewpoints (both azimuth and elevation) to handle viewpoint variations in real data. 
Finally, to provide a seamless break from the original image, an object segmentation module is desirable so that the missing pixels in background can be separately filled in by alternative methods, such as patch-based inpainting methods~\cite{barnes_siggraph2009} or pixel-wise autoregressive models~\cite{van2016pixel}.
%Finally, to deal with background disocclusion, an object segmentation module is needed so that the missing pixels in background can be filled in separately by alternative methods, such as patch-based inpainting methods~\cite{OM3D2014} or recurrent neural networks~\cite{van2016pixel}.

\section{Conclusion and Future Work}
We present a novel transformation-grounded image generation network. Our method generates realistic images and outperforms existing techniques for novel 3D view synthesis on standard datasets of CG renderings where ground truth is known. Our synthesized images are even accurate enough to perform multi-view 3D reconstruction.  We further show successful results for real photographs collected from the web, demonstrating that the technique is robust.

We observed that some structures in the generated novel views, such as headlights and wheels of cars, would consistently resemble common base shapes. This is more apparent if such structures are not observed in the input view. We believe the reason is the inherently deterministic nature of our encoder-decoder architecture, which can be alleviated by incorporating approaches like explicit diverse training~\cite{lee_nips2016} or probabilistic generative modeling~\cite{Xue_nips2016,Yan_eccv2016,mizra_arxiv2014,walker_eccv2016}.

We hope that the proposed image generation pipeline might potentially help other applications, such as video prediction. Instead of pure generation demonstrated by recent approaches~\cite{mathieu_iclr2016,Vondrick_nips2016}, our approach can be applied such that each frame uses a transformed set of pixels from the previous frame\cite{walker_eccv2016,dfn_NIPS2016,finn_NIPS2016} where missing pixels are completed and refined by a disocclusion aware completion network, where disocclusion can be learned from motion estimation~\cite{walker_eccv2016,finn_NIPS2016}.

\section*{Acknowledgement}
This work was started as an internship project at Adobe Research and continued at UNC. We would like to thank Weilin Sun, Guilin Liu, True Price, and Dinghuang Ji for helpful discussions. We thank NVIDIA for providing GPUs and acknowledge support from NSF 1452851, 1526367.

{\small
\bibliographystyle{ieee}
\bibliography{egbib}
}

\newpage
\clearpage
\appendix

\section*{Appendix}

\section{Detailed Network Architectures}
\begin{figure*}[t]
\begin{center}
\includegraphics[width=\linewidth]{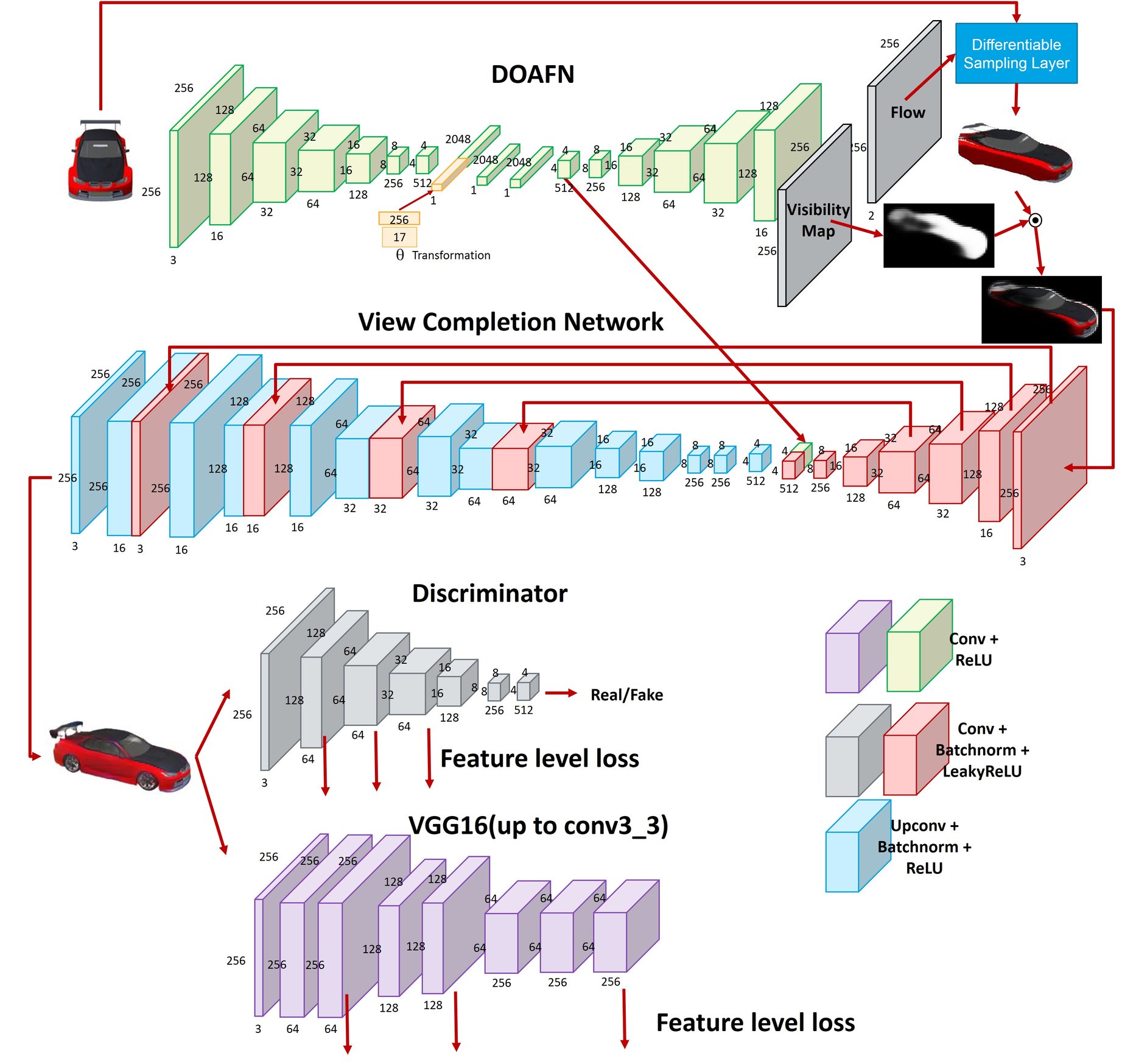}
\end{center}
\caption{Transformation-grounded view synthesis network architecture}
\label{fig:architecture}
\end{figure*}

We provide the detailed network architecture of our approach in Figure~\ref{fig:architecture}.

\section{More examples}
\begin{figure*}[t]
\begin{center}
\includegraphics[width=\linewidth]{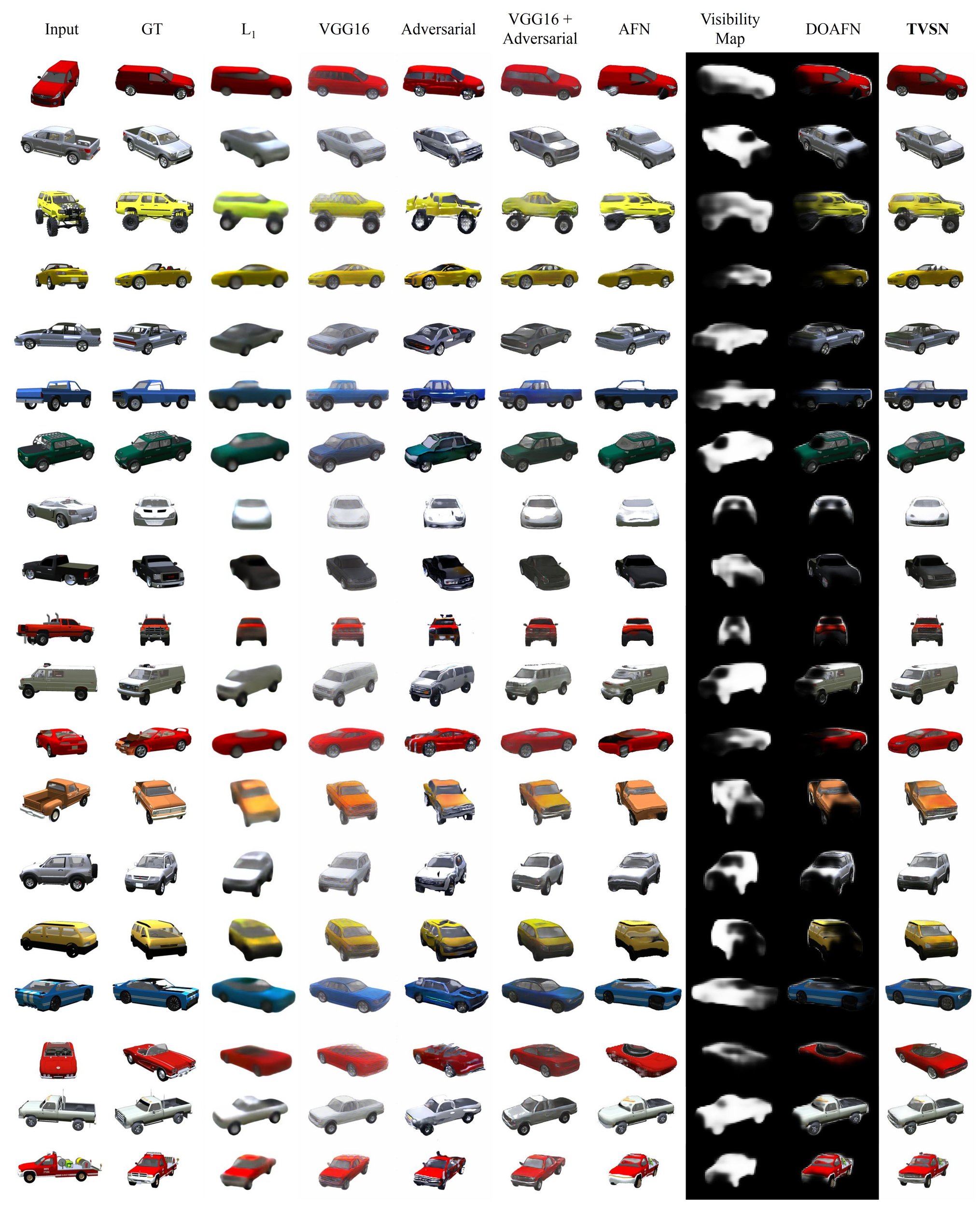}
\end{center}
\caption{Results on test images from the car category~\cite{shapenet2015}. \nth{1}-input, \nth{2}-ground truth. From \nth{3} to \nth{6} are deep encoder-decoder networks with different losses. (\nth{3}-$L_1$ norm~\cite{tatarchenko_eccv2016}, \nth{4}-feature reconstruction loss with pretrained VGG16 network~\cite{johnson_eccv2016,larsen_icml2016,ulyanov_icml2016,lamb2016discriminative}, \nth{5}-adversarial loss with feature matching~\cite{Goodfellow_nips2014,Radford_iclr2016,salimans_nips2016}, \nth{6}-the combined loss). \nth{7}-appearance flow network (AFN)~\cite{Zhou_eccv2016}. \textbf{\nth{8}-ours(TVSN)}.}
\label{fig:supp_car}
\end{figure*}

\begin{figure*}[t]
\begin{center}
\includegraphics[width=\linewidth]{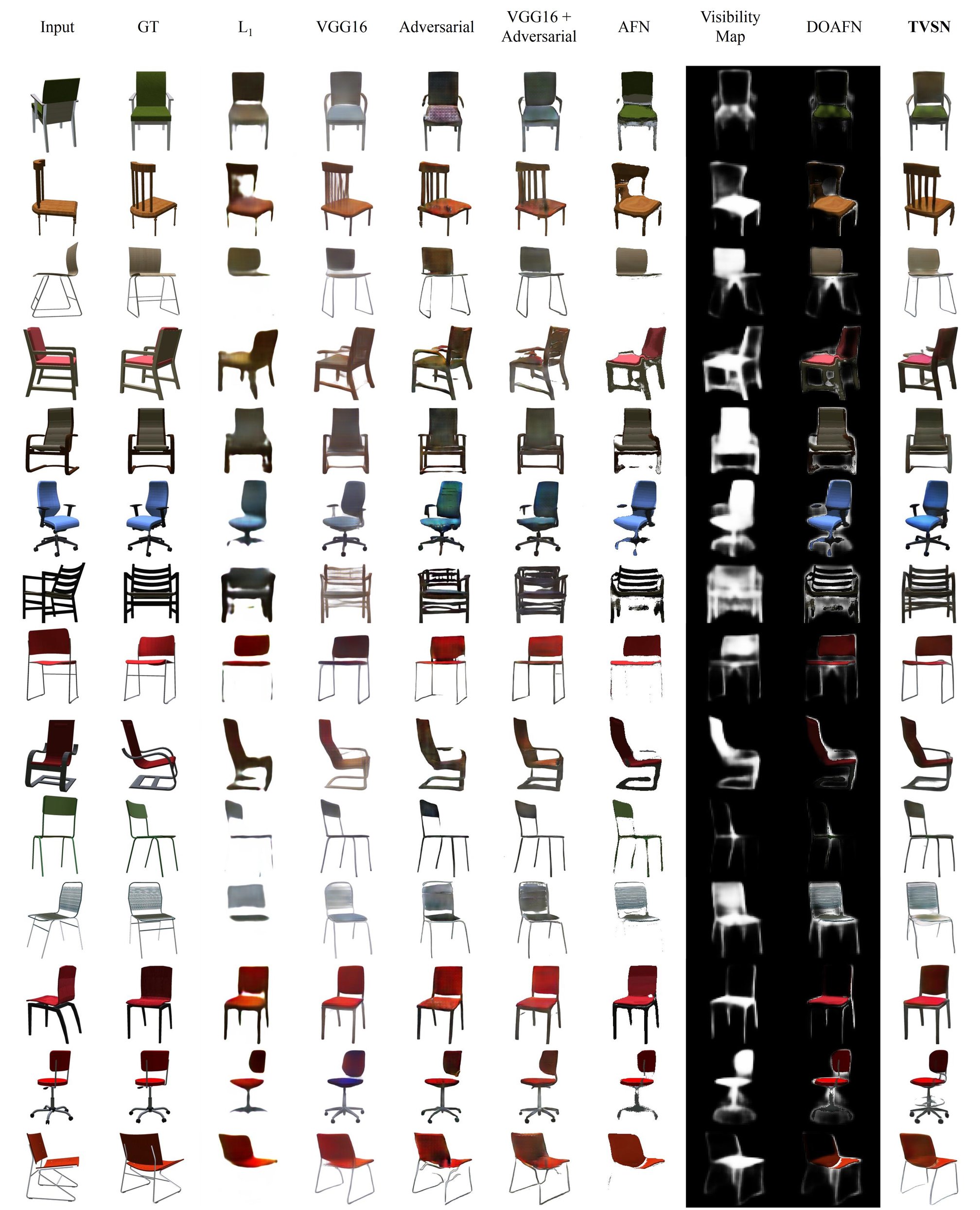}
\end{center}
\caption{Results on test images from the car category~\cite{shapenet2015}. \nth{1}-input, \nth{2}-ground truth. From \nth{3} to \nth{6} are deep encoder-decoder networks with different losses. (\nth{3}-$L_1$ norm~\cite{tatarchenko_eccv2016}, \nth{4}-feature reconstruction loss with pretrained VGG16 network~\cite{johnson_eccv2016,larsen_icml2016,ulyanov_icml2016,lamb2016discriminative}, \nth{5}-adversarial loss with feature matching~\cite{Goodfellow_nips2014,Radford_iclr2016,salimans_nips2016}, \nth{6}-the combined loss). \nth{7}-appearance flow network (AFN)~\cite{Zhou_eccv2016}. \textbf{\nth{8}-ours(TVSN)}.}
\label{fig:supp_chair}
\end{figure*}

We provide more visual examples for \emph{car} and \emph{chair} categories in Figures~\ref{fig:supp_car} and~\ref{fig:supp_chair} respectively. In addition to novel views synthesized by our method, we also provide the intermediate output (visibility map and output of DOAFN) as well as views synthesized by other approaches.

\section{Test results on random backgrounds}
\begin{figure*}[t]
\begin{center}
\includegraphics[width=\linewidth]{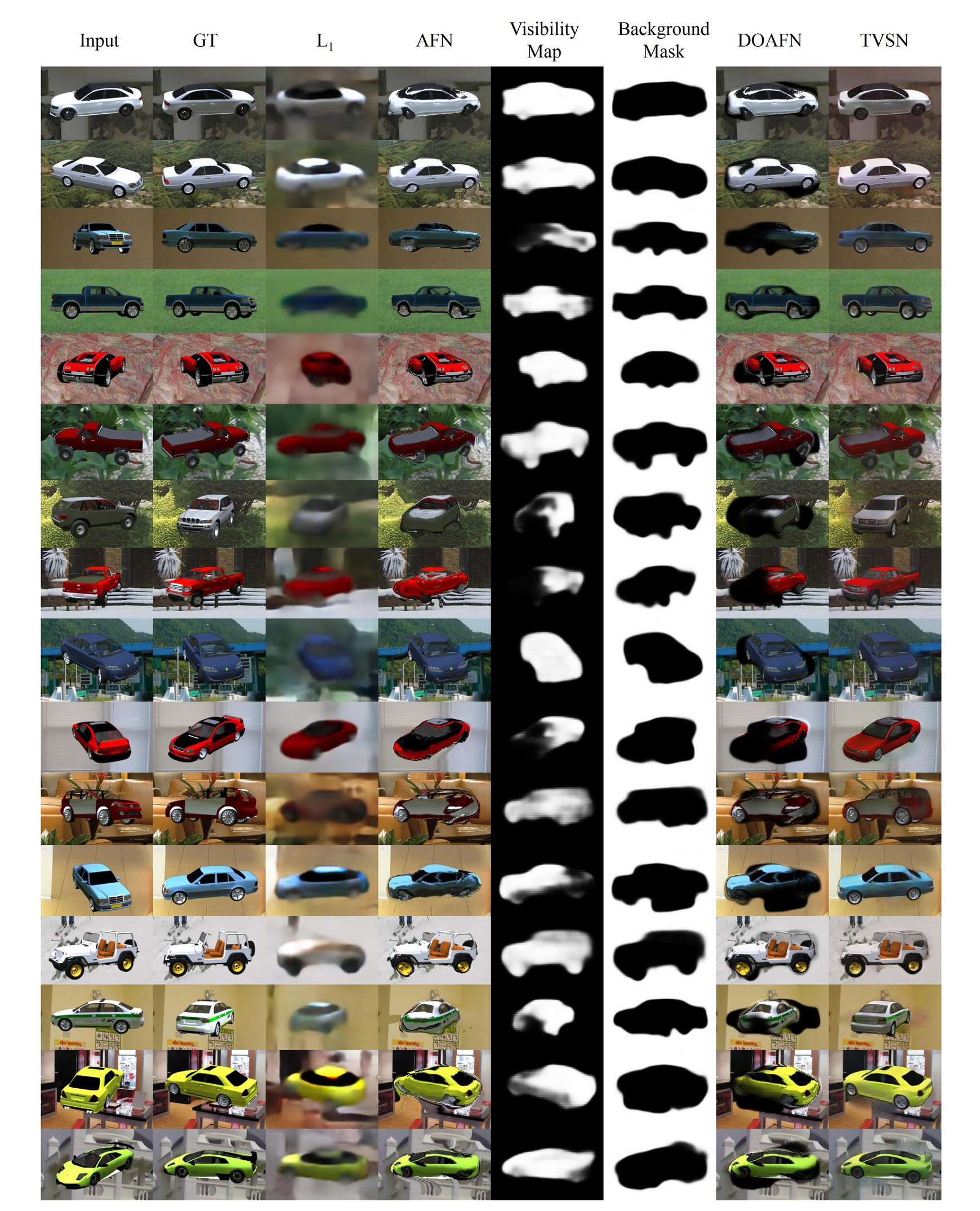}
\end{center}
\caption{Test results on synthetic backgrounds }
\label{fig:supp_bg}
\end{figure*}

Figure~\ref{fig:supp_bg} presents test results on synthesized images with random backgrounds. Intermediate stages, such as visibility map, background mask, and outputs of DOAFN are also shown. We compare against $L_1$ and AFN baselines. Note that $L_1$ and AFN could perform better on background area if we applied similar approaches used in TVSN, which we considered backgrounds separately.

\section{Arbitrary transformations with linear interpolations of one-hot vectors}
\begin{figure*}[t]
\begin{center}
\includegraphics[width=\linewidth]{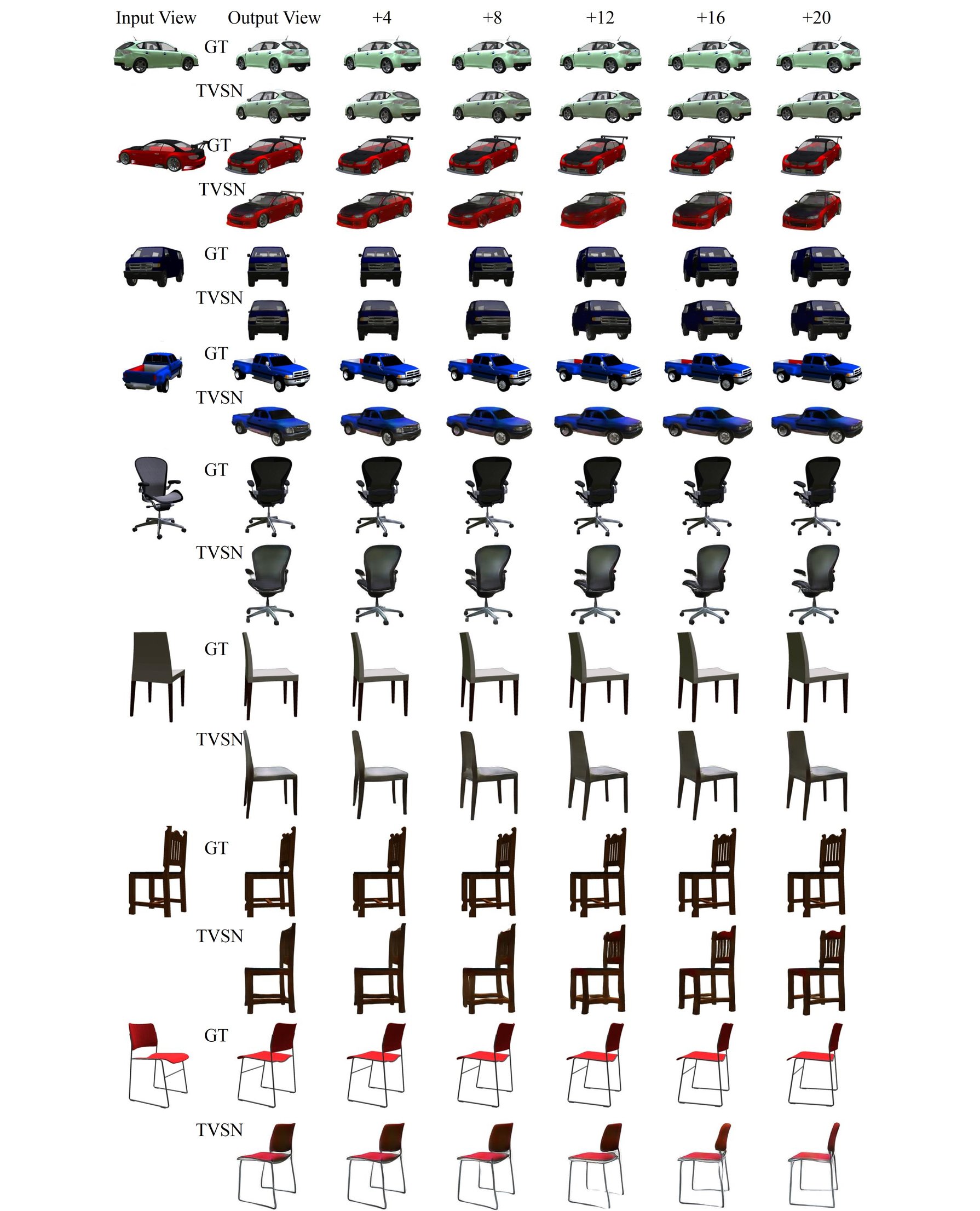}
\end{center}
\caption{Test results of linear interpolation of one-hot vectors}
\label{fig:supp_interp}
\end{figure*}

We show an experiment on the generalization capability for arbitrary transformations. Although we have trained the network with 17 discrete transformations in the range [20,340] with 20-degree increments, our trained network can synthesize arbitrary view points with linear interpolations of one-hot vectors. For example, if [0,1,0,0,...0] and [0,0,1,0,...0] represent 40 and 60-degree transformations respectively, [0,0.5,0.5,0,...0] represents 50 degree. More formally, let $\mathbf{t} \in [0,1]^{17}$ be encoding vector for the transformation parameter $\theta \in [20,340]$ and $s$ be step size ($s = 20$). For a transformation parameter $i \times s \leq \theta < (i+1) \times s$, $i$ and $i+1$ elements of the encoding vector $\mathbf{t}$ is
\begin{equation}
\mathbf{t}^i = 1 - \frac{\theta - (i \times s)}{s}, \quad \mathbf{t}^{i+1} = 1 - \mathbf{t}^i
\end{equation}
Figure~\ref{fig:supp_interp} shows some of examples. From the third to the sixth columns, we used linearly interpolated one-hot vectors to synthesize views between two consecutive discrete views that were in the original transformation set (the second and the last columns).

\section{More categories}
\begin{figure*}[t]
\begin{center}
\includegraphics[width=0.68\linewidth]{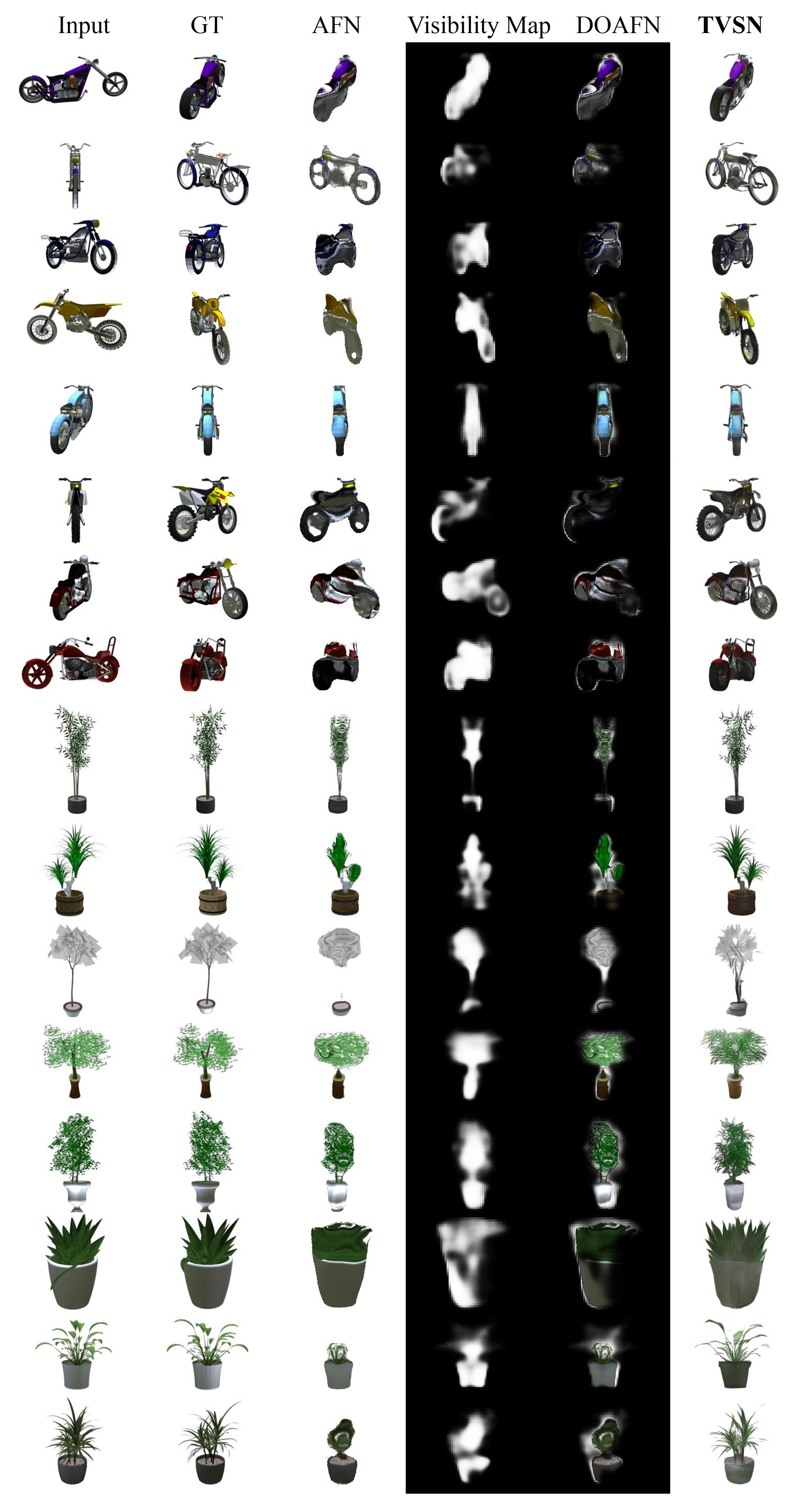}
\end{center}
\caption{Test results of motorcycle and flowerpot categories}
\label{fig:supp_others}
\end{figure*}

We picked cars and chairs, since both span a range of interesting challenges.  The car category has rich variety of reflectance and textures, various shapes, and a large number of instances. The chair category was chosen since it is a good testbed for challenging `thin shapes', e.g. legs of chairs, and unlike cars is far from convex in shape. We also wanted to compare to previous works, which were tested mostly on cars or chairs. In order to show our approach is well generalizable to other categories, we also performed experiments for \emph{motorcycle} and \emph{flowerpot} categories. We followed the same experimental setup. We used the entire \emph{motocycle}($337$ models) and \emph{flowerpot}($602$ models) categories. For each category, 80\% of 3D models are used for training, which leaves around $0.1$ million training pairs for the \emph{motorcycle} and $0.2$ million for the \emph{flowerpot} category. For testing, we randomly sampled instances, input viewpoints, and desired transformations from the rest 20\% of 3D models. Figure~\ref{fig:supp_others} shows some of qualitative results.

\end{document}